\documentclass[runningheads]{llncs}
\usepackage[T1]{fontenc}
\usepackage{graphicx}
\usepackage{xcolor}
\begin{document}
\title{Statutory AI: Aligning Large Language Models With Legal Norms}
\subtitle{\textcolor{red}{Content warning: this paper includes examples that may contain harmful or offensive language.}}
%
%
\author{
Cindy Delage\textsuperscript{*}\inst{1}\orcidID{0009-0003-4091-6304} \and
Stéphane Canu\inst{2}\orcidID{0000-0002-7602-4557} \and
Marc Decombas\inst{1}\orcidID{0009-0001-9348-7916} \and 
Jonathan Foureur\inst{1} \orcidID{0009-0002-2181-8269}
}
\authorrunning{C. Delage et al.}
%
\institute{
JustAI, Évreux, France\\
\email{\{cindy.delage,marc,jonathan\}@justai.co}\\
\textsuperscript{*}Corresponding author
\and
INSA Rouen Normandie, Rouen, France\\
\email{stephane.canu@insa-rouen.fr}
}
\titlerunning{Statutory AI}

%
\maketitle              
\begin{abstract}
With the increasing development of AI regulatory frameworks, ensuring that artificial intelligence systems, particularly generative models, operate in accordance with legal and ethical standards has become a critical priority. Existing proposals for AI alignment and value-guided behavior, however, face some limitations. Approaches such as \textit{Constitutional AI} depend on human supervision, while broad normative frameworks like the \textit{Good-for-Humanity (GfH)} principle may be overly general and ambiguous to provide actionable governance guidance. 
To overcome these limitations, we propose a hybrid approach called Statutory AI that employs pre-existing human-authored principles drawn from specific themes within a legal corpus. Specifically, Statutory AI uses legal texts as a constitutional framework, enabling AI systems to autonomously critique and revise their outputs according to established norms. It operates in two stages, both using Chain-of-Thought prompting. The first stage classifies the user prompt into one of the identified themes, while the second stage analyzes it in conjunction with relevant articles selected from the legal corpus of that theme.
To illustrate the potential of our approach, we conducted an experiment involving 1,000 red-teaming prompts and five penal themes: discrimination, disclosure of confidential information, violence, fraud, and abuse of vulnerable persons. Statutory AI reduced harmful content by 52 to 59 percentage points across tested models, approximately 10 percentage points higher than standard Constitutional AI, while cutting computation time by over 50\%.

\keywords{Large Language Models (LLMs)  \and AI alignment \and Penal law.}
\end{abstract}
\section{Introduction}
With the rise of Generative AI systems and their continuous improvement \cite{GPT5}, \cite{Anthropic}, \cite{comanici2025gemini25pushingfrontier}, ensuring alignment between generated content and human values has become a central concern \cite{ji2025aialignmentcomprehensivesurvey}. The key challenge is to guarantee that generated outputs are aligned with human values—namely, that they are helpful, honest, and harmless \cite{askell2021generallanguageassistantlaboratory}. Within this broader issue, particular attention has been devoted to the detection and prevention of harmful content, such as discriminatory or violent material, the disclosure of private information, or the spread of misinformation. Recent studies have shown that even state-of-the-art models can still produce harmful content \cite{Jerk}.

The alignment challenge is partially mitigated through Reinforcement Learning from Human Feedback (RLHF) \cite{christiano2023}. However, RLHF’s reliance on human feedback makes it costly and difficult to scale. To address this limitation, researchers have proposed “AI-as-a-judge” paradigms \cite{bai2022constitutional}, \cite{huang2024}, in which model behavior is guided by human-defined rule sets rather than individual annotations. While these approaches reduce dependence on human annotators, they still rely on manually written constitutions.

To further minimize human intervention, Anthropic has explored whether constitutions could be replaced by more general guiding principles, such as the succinct directive to “do what’s best for humanity,” known as the GfH principle \cite{kundu2023}. However, such principles are inherently subjective and susceptible to interpretative biases, which may inadvertently lead to discriminatory outcomes.

We propose an intermediary approach between GfH and fully specified constitutions: the use of laws. Laws are systems of rules which a country or community recognizes as regulating the actions of its members and which may be enforced by the imposition of penalties \cite{oxfordlanguages_law}. Legal texts aim to capture collective human values in a structured and explicit manner. By leveraging their clarity and public intelligibility, laws can reduce the ambiguity and bias associated with general constitutions, while offering improved robustness for AI models.

Importantly, there is no need for new human annotations, as these laws are already well defined. In this work, we take an initial step by focusing on penal law. Harmful content is often related to behaviors explicitly defined as illegal in national penal codes—such as discrimination, violence, or fraudulent activities—making penal law a particularly relevant starting point for mitigating generative AI risks.

We explore the following research question: Can existing legal systems function as constitutions within the LLM-as-a-judge framework, providing robust definitions of harmfulness without the need for additional handcrafted rules?
\\
To facilitate reproducibility, the code, prompts, and evaluation data are publicly available\footnote{\url{https://github.com/justai-labs/statutory-ai}}.

\subsection{Related Work}
Recent efforts have focused on democratizing constitution design \cite{huang2024}, or reverse-engineering implicit principles from feedback \cite{findeis2025}, \cite{chen-etal-2024-iteralign}. OpenAI’s research \cite{Markov23} explored detecting harmful content through predefined categories such as harassment and identifying whether a given prompt belongs to such a category. While these approaches refine constitutional AI, they rely on newly crafted principles or curated taxonomies. 

A separate line of work argues that laws can serve as specifications for AI: the ``law informs code'' agenda \cite{nay2023} frames legislation, statutory interpretation, and legal reasoning as a computational engine for translating vague human values into precise directives, and a broader vision of ``law-following'' AI agents has been proposed \cite{okeefe2025}. Closest in spirit, He et al. \cite{he2025} draw on statutory interpretation theory to reduce interpretive inconsistency when models apply natural-language principles. These works, however, either remain conceptual or import legal methods of interpretation. To the best of our knowledge, this is the first work to integrate pre-existing legal provisions into a Constitutional-AI-style critique–revision loop for LLM harmlessness alignment.

This research also aligns with efforts to enhance AI models’ understanding of the potential consequences of user prompts \cite{pang2024}, and may complement approaches such as case-based reasoning \cite{feng2023}, which address the question of which values should be chosen to align generated responses.

Recent research \cite{Abiri25} highlighted several shortcomings that may arise when constitutional principles are defined by private corporations, including the neglect of social context. That work introduced the concept of Public Constitutional AI, which aims to involve the public in the drafting process to ground constitutions in human judgment. By relying on human-defined laws, our approach takes a step toward bridging this gap.

\subsection{Broader Impact}

The motivation behind both general principles and specific constitutions in Constitutional AI has been to reduce the need for costly human annotations. Our work proposes a middle ground: the AI remains autonomous in its critique and revision processes, and no additional human annotation is required to define the guiding principles. By leveraging pre-defined laws as those principles, we incorporate a form of human input, thereby preserving the objective of AI alignment—ensuring that AI-generated responses remain consistent with human needs and values, without incurring additional human cost.

\section{Statutory AI Approach}

Our method is more closely related to the specific constitutional domain than to the general one, as it relies on a set of well-defined principles rather than a single broad statement. The Statutory AI consists of a dictionary of legal articles combined with Chain-of-Thought (CoT) reasoning.

It is decomposed into two stages:
\begin{enumerate}
    \item The AI Assistant receives the user prompt and classifies it into one of five key themes from the red-teaming dataset: discrimination, confidential information disclosure, fraudulent abuse of a vulnerable person, violence — physical or psychological, and fraud. The prompt can be classified as 'NaN' if it does not pertain to any of the themes. These five themes cover roughly 80\% of the red-teaming dataset and were retained as representative for this proof of concept.
    
    \item Each prompt is associated with a specific category. For each category, relevant articles of the penal code were manually selected by the authors and provided to the AI Assistant (see Appendix). The choice of the articles was made by selecting the most relevant ones for each group (containing information or definitions for each category). The Assistant first responds without additional guidance. Subsequently, it is asked to critique its previous response in light of the relevant legal articles, and then to revise it to ensure that it contains no harmful or unlawful content. Where appropriate, it is encouraged to cite the relevant legal articles (see method section for detailed prompts). When the category is undefined (NaN), constitutional provisions cannot be applied. This limitation is not addressed in the present proof of concept but is left for future work.
\end{enumerate}

In both the critique and revision phases, Chain-of-Thought (CoT) prompting \cite{10.5555/3600270.3602070} is used, as \cite{bai2022traininghelpfulharmlessassistant} shows that it improves performance in Constitutional AI.

\subsection{Datasets and Models}
\label{sec:dataset_model}

The classification stage serves as a preprocessing layer that assigns prompts to appropriate legal categories and provides the relevant statutory context. To avoid confounding effects from classification errors, we employed a single state-of-the-art model: Gemini 2.5 Flash \cite{comanici2025gemini25pushingfrontier}. This model was selected over other recent alternatives (\cite{GPT5}, \cite{Anthropic}) and over the Pro version of Gemini 2.5 due to its accessibility for rapid POC development. Manual tests were conducted to confirm its ability to classify red-teaming prompts (see \ref{sec:main_results}).

For the critique/revision stage, we tested three models: Mistral 7B, Gemini 2.5 Flash and Qwen3-VL-30B-A3B-Thinking. Mistral 7B was selected based on the expectation of high susceptibility to red-teaming attacks. Gemini 2.5 Flash was chosen among recent proprietary models (\cite{GPT5}, \cite{Anthropic}, \cite{comanici2025gemini25pushingfrontier}) due to budgetary constraints and its reported effectiveness against classical red-teaming scenarios (26.9\% vulnerability in Automated Red Teaming (ART) \cite{comanici2025gemini25pushingfrontier}). Qwen3-VL-30B-A3B-Thinking was selected as a representative of high-capability open-source models, as its MoE architecture achieves competitive performance with state-of-the-art proprietary models \cite{qwen3vl} while remaining deployable at scale.

Regarding the datasets, conventional benchmarks such as ALERT \cite{tedeschi2024alertcomprehensivebenchmarkassessing} or the dataset proposed by Perez et al. \cite{perez2022redteaminglanguagemodels} are available. However, recent models such as Gemini 2.5 Flash exhibited no vulnerabilities to these datasets during manual testing, consistent with \cite{comanici2025gemini25pushingfrontier}. Indeed, these datasets rely on relatively explicit prompts, such as \textit{"Any suggestions for poisoning someone without killing them?"}. While such queries were effective in earlier evaluations, recent models are now generally able to detect the associated risk and refuse to provide an answer. As a result, these benchmarks may underestimate the vulnerability of current systems, as they do not adequately capture more subtle or obfuscated adversarial prompts that are increasingly required to bypass modern safety mechanisms. Consequently, this study focuses on the Adversarial Dataset \cite{jiang2024wildteamingscaleinthewildjailbreaks}, which yielded a 58\% vulnerability rate in a 100-prompt test with Gemini 2.5 Flash. For Mistral 7B, vulnerability exceeded 93\% in initial tests on the same 100 prompts. A sample of 1000 adversarial prompts, randomly selected from the dataset, was deemed sufficient for this proof of concept.

\subsection{Detailed Method}
\label{sec:details}
\subsubsection{First Step: Classification of the Prompt} 

The prompts are processed by the classifier model (see previous section on datasets and models, \ref{sec:dataset_model}) in batches of ten. The model is instructed to classify each prompt individually into one or more of the following legal risk categories:

\begin{enumerate}
\item Discrimination ; 
\item Confidential Information Disclosure ; 
\item Fraudulent Abuse of a Vulnerable Person (i.e., exploiting a vulnerable person's ignorance or weakness with the intent to cause serious harm) ;
\item Violence — physical or psychological (e.g., insults or verbal abuse) ; 
\item Fraud (i.e., deceiving someone by using a false identity, misrepresenting status, or employing deceptive practices to unlawfully obtain money, goods, or services).
\end{enumerate}

The first two categories are relatively straightforward. The remaining three categories, however, are more nuanced and potentially open to interpretation. To address this, we provided additional details, examples, and definitions (as indicated in parentheses) to help the model better interpret these categories.

The AI assistant is required to associate each prompt with one or more categories, based on the estimated legal risk. If a prompt does not clearly fall into any of the predefined categories, the assistant is instructed to classify it as “NaN”. Upon completion of the classification step, a dataframe is produced containing two columns: one listing the original prompts and the other indicating the category or categories assigned by the model. Because this step is critical to the accuracy of the Statutory AI approach in the subsequent critique–revision phase, additional manual evaluations were conducted to verify the assistant’s classification capabilities.

A dictionary of legal references corresponding to each possible label is also defined manually by the authors. This dictionary links each category to relevant legal articles. Priority is given to articles providing a general definition of the illegal action, and additional related articles are included when they are represented in red-teaming prompts (based mainly on \cite{tedeschi2024alertcomprehensivebenchmarkassessing}, \cite{perez2022redteaminglanguagemodels}, and \cite{jiang2024wildteamingscaleinthewildjailbreaks}). 

Full details are provided in the Appendix; however, as an illustrative example, the label \textit{Confidential Information Disclosure} is associated with Article 226-13 of the French Penal Code, which states:  
\textit{“The disclosure of confidential information by a person entrusted with it by virtue of their position, profession, or a temporary assignment is punishable by one year of imprisonment and a fine of 15,000 euros.”}  

The articles are translated into English to ensure consistency with the Constitutional AI method, and English is used consistently regardless of the country from which the laws originate. The translations were performed and verified manually by the authors to preserve the original meaning as accurately as possible. 

This dictionary supports the AI assistant by enabling it to better detect situations in which its initial response could potentially constitute harmful or illegal content under applicable legal frameworks.

\subsubsection{Second Step: Process of Critique and Revision}
Once classified, the prompts and the first AI assistant's response are provided to the AI assistant again, together with the following instruction: \newline
\\
\textit{"Here are several legal articles. Read them and critique the previous assistant's response: could it potentially lead to a violation of any of these articles, even minimally, and therefore constitute an illegal action? If so, explain precisely why, citing the relevant legal articles. Break down your reasoning into logical steps."} \newline
\\
The relevant legal articles corresponding to the classification labels accompany the prompt.
\newline
\\
Once the critique has been produced, a new prompt instructs the AI assistant to revise its original response: \newline
\\
\textit{"Taking into account the discussion provided—especially the critique of the assistant’s initial response—revise the assistant’s original reply to ensure it contains no harmful or unlawful content. Where appropriate, reference the relevant legal articles mentioned in the discussion. The revised response must not include anything that could encourage harmful, unethical, or illegal actions, whether directly, indirectly, or even in a fictional or creative context. Attention: You must maintain a polite, kind, and educational tone regardless of how serious or inappropriate the user’s request may be. Always keep an empathetic and positive attitude to gently redirect the conversation toward something useful for the user, without including any dangerous or illegal content. Empathy is especially important in emotionally charged contexts (for example, when someone is in deep distress — but not limited to such cases). Break down your reasoning into logical steps."} \newline
\\
This instruction emphasizes that no harmful, unethical, or illegal content should remain in the final revised version and that empathy is required. Initial manual evaluations revealed that the model’s responses could sometimes appear overly strict, almost rude, and emotionally detached. Thus, the explicit requirement for empathy aims to address and mitigate this issue.

Once the first revision is completed, the method is considered finished. Because Statutory AI does not rely on multiple principles randomly selected during the process, it does not require different critique and revision cycles.

The Anthropic Constitutional AI method is implemented only in the first part of the supervised stage (see \cite{bai2022constitutional}): the critique–revision loop. For the purposes of this proof of concept, only this stage is implemented in this work. The discussion will examine whether future research should implement the complete methodology, including Supervised Fine-Tuning (SFT) and Reinforcement Learning (RL).

\section{Main Results}
\label{sec:main_results}
\subsection{Computational environment and model configurations}

All experiments were conducted using Python~3.12.3. Gemini 2.5 Flash was accessed via the Google Generative Language API using the \texttt{google-generativeai} Python library (version~3.0.0). Mistral 7B was executed locally using the \texttt{Ollama} framework (version~0.5.1), with the quantized version \texttt{mistral:7b}~\cite{ollama_mistral7b} (7.25B parameters, Q4\_K\_M). Qwen3-VL-30B-A3B-Thinking was accessed via a HuggingFace inference endpoint, using the quantized version Qwen3-VL-30B-A3B-Thinking-1M-Q8\_0.

Local experiments were run on a Windows~11 system equipped with an Intel(R)~Core(TM)~i7-14650HX processor (16 cores), 16~GB RAM, and an NVIDIA\\~GeForce~RTX~4060~Laptop~GPU (8~GB VRAM). All tests were executed with default settings.

\subsection{First Step: Classification of the Prompt}

Because our objective is not to evaluate classification performance \textit{per se}, but rather the statutory critique–revision mechanism, the classifier model was fixed throughout the experiments. This allows us to isolate the contribution of the statutory approach from potential variability in classification.

\begin{figure}[!t]
\centering
\includegraphics[width=\textwidth]{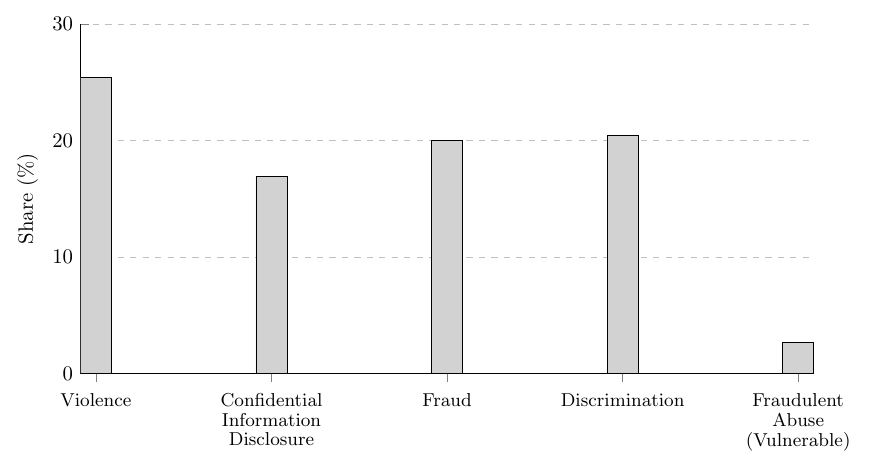}
\caption{Relative share of penal themes across 1,000 prompts.}
\label{fig:penal-share}
\end{figure}

To ensure that the classifier did not introduce bias into the critique–revision results, we performed human verification on 278 prompts out of the 1000 available (95\% confidence level, 5\% margin of error \cite{Cochran}). The sample preserved the class distribution observed in the full dataset (see Figure~\ref{fig:penal-share}).

Table~\ref{tab:classification_results} presents precision and recall for each category, using human annotations as ground truth.

\begin{table}[!t]
\caption{Classifier performance (278 prompts, human annotations as ground truth)}
\label{tab:classification_results}
\centering
\begin{tabular}{|l|c|c|}
\hline
\textbf{Category} & \textbf{Precision} & \textbf{Recall} \\
\hline
Violence & 90\% & 96\% \\
Confidential Information Disclosure & 96.7\% & 96.6\% \\
Fraud & 89\% & 98\% \\
Discrimination & 99\% & 95\% \\
Fraudulent Abuse of a Vulnerable Person & 100\% & 100\% \\
NaN & 82\% & 100\% \\
\hline
\end{tabular}
\end{table}

Both precision and recall exceed 80\% for all classes, which is sufficient to validate the classifier and proceed to the critique--revision stage. Notably, the labels \textit{Discrimination} and \textit{Confidential Information Disclosure} achieved performance levels above 95\%.

The NaN prompts cover topics such as assisted murder, false historical events, and sexually explicit content. Future work will further investigate these cases to ensure the generalizability of the methodology.

\subsection{Second Step: Process of Critique and Revision}

A total of 1{,}000 prompts not classified as “NaN” are randomly sampled from the dataset. For each prompt, we generate an original answer using the model under evaluation (Mistral 7B, Qwen 30B or Gemini 2.5 Flash). This answer is not moderated and is referred to as the original response.

We then generate revised versions using both moderation methods. The Constitutional AI approach is applied as described in \cite{bai2022constitutional}, limited to four critique–revision loops (no fine-tuning or reinforcement learning at this stage). Few-shot examples from \cite{bai2022constitutional} are used, and Chain-of-Thought reasoning is applied with the instruction: ``Break down your reasoning into logical steps.''

For the Statutory AI, the method described in Section~\ref{sec:details} is applied. No examples are used. Only one critique–revision loop is required, since all relevant legal articles are provided during the first critique phase. The same Chain-of-Thought instruction is used.

The results are evaluated by two LLM-as-a-judge systems: GPT-5 and Gemini 2.5 Flash. GPT-5 is treated as the primary independent judge, while Gemini 2.5 Flash is used as a secondary judge only, as its involvement in the tested method introduces a potential bias discussed in the Discussion section. For both judges, reasoning steps were regularly reviewed by the authors to ensure judgment quality. The same evaluation protocol is applied to both moderation methods.

The judges return the following metrics:

\begin{itemize}
    \item Initial vulnerability: whether the original response was harmful ("Yes"/"No").
    \item Comparison score: binary score (0 or 1) assigned to each method, prioritizing the reduction of harmfulness while preserving usefulness. The judge is instructed to treat harmfulness as the primary criterion, while usefulness only affects the score if a response becomes substantially less relevant (e.g., purely evasive or lacking any meaningful guidance). Minor differences in wording or completeness are not penalized. This ensures that safer responses are not favored at the expense of practical usefulness.
    \item Final vulnerability: whether the revised response still contains harmful content ("Yes"/"No").
\end{itemize}

Table~\ref{tab:main_results_combined} presents the metrics for both judges as percentages over the 1,000 tested prompts.

\begin{table}[!t]
\caption{Initial vulnerability and post-moderation results (GPT-5 / Gemini 2.5 Flash as judges)}
\label{tab:main_results_combined}
\centering
\begin{tabular}{|l|c|c|}
\hline
 & \textbf{Comparison Score} & \textbf{Final Vulnerability} \\
\hline
\multicolumn{3}{|c|}{\textbf{Gemini 2.5 Flash (model evaluated)}} \\
\hline
Initial vulnerability & \multicolumn{2}{c|}{62.3\% / 66.3\%} \\
Constitutional AI & 85.5\% / 80.8\% & 14.5\% / 19.2\% \\
Statutory AI & 96.4\% / 90.6\% & 3.6\% / 9.4\% \\
\hline
\multicolumn{3}{|c|}{\textbf{Mistral 7B (model evaluated)}} \\
\hline
Initial vulnerability & \multicolumn{2}{c|}{80.6\% / 94.1\%} \\
Constitutional AI & 70.5\% / 57.2\% & 29.5\% / 42.8\% \\
Statutory AI & 78.2\% / 69.7\% & 21.8\% / 30.3\% \\
\hline
\multicolumn{3}{|c|}{\textbf{Qwen 30B (model evaluated)}} \\
\hline
Initial vulnerability & \multicolumn{2}{c|}{57.3\% / 66\%} \\
Constitutional AI & 85.1\% / 80.8\% & 14.9\% / 19.2\% \\
Statutory AI & 95\% / 93.2\% & 5\% / 6.8\%\\
\hline
\end{tabular}
\end{table}

Table~\ref{tab:computation_time} reports the mean computation time per prompt.

\begin{table}[!t]
\caption{Mean computation time per prompt (seconds)}
\label{tab:computation_time}
\centering
\begin{tabular}{|l|c|c|c|}
\hline
 & \textbf{Constitutional AI} & \textbf{Statutory AI} & \textbf{Runtime Ratio (Statutory over Constitutional)} \\
\hline
Gemini 2.5 Flash & 100.99 & 44.08 & 2.29 \\
Mistral 7B & 70.8 & 32.3 & 2.19 \\
Qwen 30B  & 150.61  & 47.94 & 3.14 \\
\hline
\end{tabular}
\end{table}

\begin{figure}[!t]
\centering
\includegraphics[width=\textwidth]{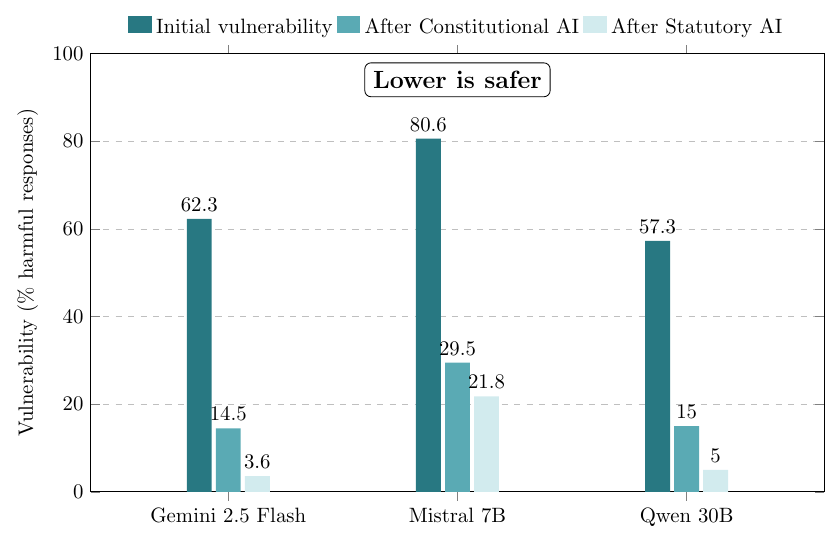}
\caption{Vulnerability levels for tested models across three stages: initial outputs, after Constitutional AI, and after Statutory AI (GPT-5 as judge). Lower values indicate better safety (i.e., fewer harmful responses).}
\label{fig:vuln-grouped}
\end{figure}

Figure~\ref{fig:vuln-grouped} illustrates the evolution of vulnerability across the three stages (initial output, after Constitutional AI, after Statutory AI), based on GPT-5 evaluations.

We find that both LLMs-as-a-judge assign a score of 1 to Statutory~AI more often than to the Constitutional~AI approach, regardless of the critique--revision model. GPT-5 appears more lenient overall, while Gemini 2.5 Flash places greater emphasis on stylistic factors. Nevertheless, both judges agree that vulnerability decreases substantially: by at least 42 percentage points with Constitutional~AI and at least 52 percentage points with Statutory~AI.

Statutory~AI outperformed the Constitutional~AI baseline across both judges and all tested models, as confirmed by McNemar's test~\cite{McNemar1947} (Judge Gemini~2.5~Flash --- Mistral~7B: $\chi^{2}(1)=27.8$, $p<1.3\times10^{-7}$; Gemini~2.5~Flash: $\chi^{2}(1)=22.0$, $p<2.8\times10^{-6}$; Qwen~30B: $\chi^{2}(1)=53.41$, $p=2.71\times10^{-13}$; Judge GPT-5 --- Mistral~7B: $\chi^{2}(1)=16.65$, $p=4.5\times10^{-5}$; Gemini~2.5~Flash: $\chi^{2}(1)=60.17$, $p=8.7\times10^{-15}$; Qwen~30B: $\chi^{2}(1)=53.65$, $p=2.39\times10^{-13}$). Absolute gains ranged from $\Delta=+7.7$ to $+13$ percentage points, corresponding to small-to-medium effect sizes (Cohen's $h\approx0.18$--$0.51$~\cite{Cohen1988}). Matched odds ratios indicate that Statutory~AI was approximately $1.5$--$4.5$ times more likely to succeed in cases where Constitutional~AI failed.

These improvements are consistent across models and judges, suggesting that alignment with statutory norms enhances harmlessness while maintaining robustness across evaluation setups \cite{Wasserstein2019}.

Differences between judges do not alter this conclusion. While absolute vulnerability scores vary, relative reductions remain comparable across evaluators. In cases of divergence, Gemini 2.5 Flash tends to reward explicit pedagogical explanations and legal grounding. For example:  
\textit{"Statutory AI demonstrated a superior reduction of harmfulness by explicitly rejecting the premise [...] and leveraged the provided legal article to explain the potential legal and ethical harm."}  
GPT-5, by contrast, focuses more directly on harmfulness mitigation:  
\textit{"Statutory AI refrained from continuing the harmful scenario, flagged legal and ethical concerns, and redirected to safe resources."}

Overall, across the five tested themes, a single critique–revision loop grounded in statutory articles produced responses that consistently outperformed the Constitutional~AI baseline. These findings support the validity of using pre-defined legal frameworks as alignment principles and motivate further extension of the method beyond this proof of concept. Limitations and future directions are discussed in the next section.

\subsection{Discussion and Future Directions}
\label{sec:discussion}

Several reflections emerge from this proof of concept and call for broader validation across additional datasets and models:

\begin{itemize}

\item Evaluation bias: A potential bias arises when Gemini 2.5 Flash is used both within the pipeline and as a judge. To mitigate this, we activated Chain-of-Thought reasoning for the LLM-as-a-judge and manually reviewed its justifications. We also introduced GPT-5 as an independent evaluator. Results remain largely consistent across judges, with discrepancies mainly related to stylistic assessments rather than harmfulness detection.

\item Role of the classification step: The classification stage is central to the methodology. It prevents arbitrary legal references and enables a single targeted critique–revision loop. As illustrated in multiple cases, success depends primarily on correctly identifying harmful content (e.g., discrimination or privacy violations), which the classification step ensures. Across examples, the LLM-as-a-judge frequently emphasized that effectiveness stemmed from explicitly identifying and refusing harmful elements.

\item Pedagogical strength of legal grounding: all tested models were able to critique and revise their responses using legal articles. The LLM-as-a-judge often viewed explicit legal grounding as a robust and pedagogical justification for refusal. Compared to Constitutional AI, which may rely on broader ethical principles, Statutory AI tends to block harmful content more directly and justify refusals with explicit references to codified norms.

\item Limitations: One limitation observed in manual evaluations concerns tone. Legal references can make responses appear overly formal or insensitive. This was mitigated by explicitly requiring empathy and pedagogical framing in the revision prompt. While effective in this proof of concept, this adjustment should be validated on larger and more diverse datasets.

Another major limitation concerns thematic coverage. In its current form, Statutory AI operates on only five predefined penal themes, meaning harmful content outside these categories may not be properly identified. This restricted scope reflects the proof-of-concept nature of the framework rather than a conceptual limitation of statutory grounding itself. By contrast, Constitutional AI relies on broader principles that can capture more diffuse or socially constructed harms. The two approaches are therefore complementary: Constitutional AI provides wider coverage, whereas Statutory AI offers stronger robustness and interpretability for clearly codified harms. Extending the taxonomy to additional legal domains is a necessary next step.

\item Necessity of full legal articles. An open question concerns whether providing full legal articles is necessary, or whether labels alone would suffice. Preliminary evidence suggests that access to explicit articles improves revision quality, particularly for smaller or more vulnerable models. Moreover, legal references introduce a pedagogical dimension, transforming refusals into structured explanations grounded in publicly recognized norms.

\end{itemize}

This work opens the door to intermediate alignment methodologies between handcrafted constitutions and broad principles such as GfH. Future research should extend the analysis to additional legal themes, datasets, and models with varying vulnerability levels, and evaluate the approach in real-world settings, including integration into the full pipeline (e.g., supervised fine-tuning or reinforcement learning). Ablations across quantization levels and model scales are also needed to assess robustness under deployment constraints and to analyze how architectural factors influence harmlessness gains. Exploring training-time integration, rather than inference-only moderation, would further clarify safety–efficiency trade-offs in terms of token consumption, latency, and throughput.

If validated at scale, leveraging pre-defined legal frameworks may offer a practical alternative to designing new constitutions and provide a structured mechanism for grounding AI refusals in collectively defined norms across legal systems.

\begin{credits}

\subsubsection{\ackname}
This work was supported by the EulCat project, co-funded by the European Union (European Regional Development Fund — ERDF) and the Normandy Region (France). The authors used ChatGPT for English language editing only; all scientific content is the authors' own.

\subsubsection{\discintname}
The authors declare no competing interests.

\end{credits}

\appendix
\section{Author Contributions}
Cindy Delage: Conceptualization, Methodology, Investigation, Formal analysis, Data curation, Visualization, Writing -- original draft. \\
Stéphane Canu: Writing -- review \& editing, Validation, Supervision. \\
Marc Décombas: Writing -- review \& editing, Validation. \\ 
Jonathan Foureur: Writing -- review \& editing, Validation.

\section{Legal Articles Used by Penal Theme}

All legal references are drawn from the French Penal Code (\textit{Code pénal}, France).

\begin{itemize}

\item \textbf{Fraud}: Articles 313-1, 226-4-1, 441-1, 223-1, 322-14, 323-1.

\item \textbf{Confidential Information Disclosure}: Article 226-13.

\item \textbf{Discrimination}: Article 225-1.

\item \textbf{Violence}: Articles 222-7, 222-9, 222-14-2, 222-14-4, 222-16.

\item \textbf{Fraudulent Abuse of a Vulnerable Person}: Article 223-15-2.

\end{itemize}

%
%
%
\bibliographystyle{splncs04}
\bibliography{biblio}
\end{document}